# Face shape classification using Inception v3

Adonis Emmanuel DC. Tio

Electrical and Electronics Engineering Institute
University of the Philippines Diliman
Quezon City, Philippines
adonis.tio@eee.upd.edu.ph

*Abstract* — In this paper, we present experimental results obtained from retraining the last layer of the Inception v3 model in classifying images of human faces into one of five basic face shapes. The accuracy of the retrained Inception v3 model was compared with that of the following classification methods that uses facial landmark distance ratios and angles as features: linear discriminant analysis (LDA), support vector machines with linear kernel (SVM-LIN), support vector machines with radial basis function kernel (SVM-RBF), artificial neural networks or multi-layer perceptron (MLP), and k-nearest neighbors (KNN). All classifiers were trained and tested using a total of 500 images of female celebrities with known face shapes collected from the Internet. Results show that training accuracy and overall accuracy ranges from 98.0% to 100% and from 84.4% to 84.8% for Inception v3 and from 50.6% to 73.0% and from 36.4% to 64.6% for the other classifiers depending on the training set size used. This result shows that the retrained Inception v3 model was able to fit the training data well and outperform the other classifiers without the need to handpick specific features to include in model training. Future work should consider expanding the labeled dataset, preferably one that can also be freely distributed to the research community, so that proper model cross-validation can be performed. As far as we know, this is the first in the literature to use convolutional neural networks in face-shape classification. The scripts are available at https://github.com/adonistio/inception-face-shape-classifier.

*Keywords — face shape classification, retraining Inception v3*

## I. Introduction

With the advent of re-trainable custom image classifiers, it has become a lot easier to make specialized image classifiers like face shape classifiers. That is, given a frontal view image of a human face, the classifier should be able to identify whether the given face is heart-, oblong-, oval-, round-, or square-shaped.

Knowledge of one's face shape is more commonly used by fashion stylists in recommending eyewear frames [1] and hairstyles [2] that accentuate or tone down facial features. For example, when choosing an eyewear, OPSM Opticians recommend that people with oblong-shaped faces consider frames with oversized frames to balance long and wide features and avoid clear rimless frames that exaggerate length and width [3]. Other similar applications include recommendation systems for hats, make-up, jewelry, and other fashion accessories. In the future, these recommendation systems can form part of a larger personal digital assistant that is linked to social media and product advertisers. Such recommendation systems could also be used to suggest virtual or cosmetic facial alterations that can further enhance one's looks. Face-shape classification schemes could also be used in facial profiling to speed up facial recognition but more abstracted profiling schemes using system-learned classes may be more useful.

Face shape classifiers that are readily accessible in the literature come as online guides, online applications, and mobile applications. A couple of peer-reviewed scientific articles are also available. Published methods on face shape classification extract pre-defined features from an image which are then used to train classifiers using the following approaches: k-nearest neighbors (KNN) [1], linear discriminant analysis (LDA) [2], support vector machines with linear kernel (SVM-LIN) [2], support vector machines with radial basis function kernel (SVM-RBF) [2], and artificial neural networks or multi-layer perceptron (MLP) [2]. While published overall accuracies look promising in the range of 64.2% to 80.0%, it would be interesting to compare these results with that of a classifier using convolutional neural networks (CNNs) which are becoming more and more popular in image classification problems. Inception v3 is one such model that uses CNNs [4]. It was originally trained to classify 1000 classes using millions of images but it can also be easily retrained for custom image classification problems. High classification accuracy for a couple of applications like apparel classification [5] motivates its use in this paper. Using CNNs in face shape classification could potentially improve classification accuracy without the need to handpick specific features to extract from the image and then use in training the model.

In this paper, we present the experimental results obtained after retraining the Inception v3 model for face shape classification. Section II discusses the face shape classification problem, available methods in face shape classification, and the motivation in using the Inception v3 image classifier in face shape classification. Section III discusses the experimental setup used to benchmark the performance of the retrained Inception v3 model against other classifiers. This includes how the dataset was gathered and labelled, how the features were extracted for use in the non-CNN classifiers, how the different classifiers were trained, and what performance metrics were used for benchmarking. Section IV presents the experimental results on training accuracy and overall classification accuracy for various training set sizes. And finally, Section V ends the paper with the conclusions and suggestions for future work.

## II. FACE SHAPE CLASSIFICATION USING INCEPTION V3

### A. The Face Shape Classification Problem

OPSM Opticians recognizes four traditional and archaic face shapes namely the round, oval, square, and heart face shapes. A study commissioned by the same group in 2014 prescribes to add five new categories namely the kite, rectangle, teardrop, heptagon, and oblong face shapes [3]. However, a survey or online literature suggests that this nine class classification scheme is yet to become mainstream. For this reason, we only focus on five basic face shapes that is often cited in online materials namely the heart, oblong, oval, round, and square face shapes. These face shapes have the following characteristics as illustrated by the sample images shown in Fig. 1:

- Heart-shaped faces are usually characterized by wide cheekbones and forehead that taper down to a narrow chin. Most heart-shaped faces also exhibit a V-shaped hairline called a widow's peak. Ashley Greene and Jennifer Love Hewitt are notable examples of celebrities with this kind of face shape.
- Oblong-shaped faces usually appear longer than it is wider. Celebrities known to have this face shape include Giselle Bundchen and Sarah Jessica Parker.
- Oval-shaped faces exhibit a curvilinear shape with a height to width ratio of about 3:2. Emma Watson and Kate Middleton are recognized to have oval faces.
- Round-shaped faces are circular that is about as long as it is wide. Selena Gomez and Ginnifer Goodwin are examples of celebrities often cited with this face shape.
- Square-shaped faces usually exhibit strong angular jaws with a broad forehead that is almost as wide as the cheekbones and the jaw line. Celebrities with a prominent square face include Olivia Wilde and Sandra Bullock.

In general, it is difficult to classify face shapes, more so using only these qualitative guidelines as basis. In fact, a total of 62 out of 117 female celebrity names with at least two cited face shapes have conflicting face shape classifications in 21 online sites that we reviewed. Training face shape classifiers with images with widely-agreed upon face shape labels can help minimize human bias and help automate the classification process at a greater speed and consistency.

### B. Face Shape Classification Methods

Face shape classifiers that are readily accessible in the literature come as online guides, online applications, and mobile applications. There are also a couple of peer-reviewed scientific articles available.

The more rudimentary approach to face shape classification involves tracing the outline of one's face in a mirror and then comparing the resulting shape to one of several pre-defined classes. An online application removes the need to use a mirror by allowing the user to superimpose one of five available face shape outlines on an uploaded picture [6]. Both approaches need the subjective judgment of the user to choose which shape best matches his or her face shape. Other online guides employ a rules-based approach requiring the user to answer a questions including whether one's face is longer than it is wider or whether one's jaws are rounded or angular [7]. While this approach may best approximate how humans determine face shape, this approach is not easily automated in a computer.

One online application takes a picture of one's face via a webcam and extracts from this the global face shape, measurement ratios, and skin, eye, and hair color to recommend eye wears for the user to try-on virtually in real-time [8]. It is not known what specific computer vision approach and classification algorithm is used. Another online application allows the user to manually trace the following four pairs of points in an uploaded picture: chin-hairline points, left-right forehead points, left-right cheekbone points, and left-right jaw bone points. The application then determines the face shape from these four measurements [9]. A face shape classifier mobile app is also available where the user needs to manually trace the outline of the face by positioning 12 contour points [10].

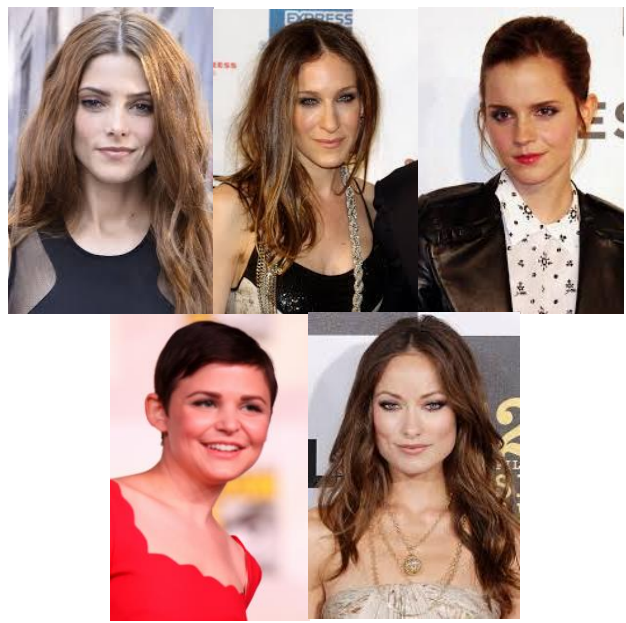

**Fig. 1** Often cited female celebrities for heart, oblong, oval, round, and square face shapes. Images are obtained from the Wikimedia Commons repository where the appropriate copyrights apply. Images shown are not representative of the training and testing images used.

Published work in face shape classification uses handpicked features computed from facial landmark coordinates to train traditional classifiers. In [1], a total of 8 features were used including the height of bounding ellipses, distance of bounding ellipses to facial boundaries, length of the jaw line, and diagonal lines from the chin to the lower ear points. The method uses a k-nearest neighbour (KNN) approach trained using 300 images. A face is said to be a blend of two face shapes if a so-called blending score falls within 40%-60%. The reported accuracy is at 80% if errors due to the failure of a sub-component that outlines the facial boundaries is included. If only cases with correctly identified facial boundaries are considered, the reported classification accuracy is about 90%.

In [2], 19 features were extracted from facial images including the face height to width ratio, jawline width to face width ratio, chin-to-mouth to jaw line width ratio, and the angles that the chin to one of 16 facial points make with respect to the horizontal. Five classifiers were trained and tested using features obtained from 500 images: linear discriminant analysis (LDA), support vector machines with linear (SVM-LIN) and radial basis function kernels (SVM-RBF), and artificial neural networks or multi-layer perceptron (MLP). Reported training accuracy for a training size of 450 images ranges from 64.9% for LDA up to 70.7% for SVM-RBF with an overall accuracy ranging from 64.2% for LDA up to 70.8%

*C. Retraining Inception v3 for Face Shape Classification*

While the reported accuracies of published work in face shape classification look promising, there is much to be desired given the recent advances in machine learning, particularly in using deep convolutional neural networks (CNNs) in image classification.

CNNs are neural networks that use convolution instead of matrix multiplication in at least one of the layers. It is especially suited for data with grid-like structures such as 1D grid audio data and 2D grid image data [11]. Typical operations inside a CNN include convolution, activation, pooling, matrix multiplication, and classification. Convolution usually reduces the dimension of the input by convolving with a kernel *K*. For a 2D image *I*, a 2D kernel *K* is often used by performing Eq. (1) to usually yield a reduced-dimension matrix if single-valued or tensor if multi-valued.

$$S(i,j) = (I * K)(i,j) = \sum_m \sum_n I(i+m, j+n) K(m,n) \quad (1)$$

Activation involves the application of non-linear operators, usually element-wise, to the resulting matrix or tensor. The more commonly used activation functions are the Rectified Linear Unit (ReLu) that returns the original input unless it is less than zero. Pooling is another common method in CNNs that further reduces the dimension by computing a summary statistic of nearby matrix or tensor values. A common pooling technique is the MaxPool which gets the maximum value within a fixed rectangular neighborhood. The resulting reduced matrix or tensor is then sometimes scaled and translated via matrix multiplication and scalar addition. Several parallel and/or cascaded convolution, activation, pooling, and affine transformation steps are performed before a final set of features can be used for classification. After the features are extracted, a softmax function can be used to choose which classes are the most probable output.

CNNs trained for image classification show excellent performance in classifying millions of images into one of 1000 classes [4]. However, due to the large number of parameters that are needed to be estimated as well as the large number of training images to be processed, usually iteratively, training a CNN from scratch may require specialized hardware and a long training time. While training a CNN image classifier from scratch may be prohibitive, experiments show that pre-trained CNNs can be retrained for specialized applications without needing the same bulk of training data and hefty computing resources. This idea is called transfer learning and is performed by training only the last few layers of the network. The motivation behind transfer learning is that the output of the first few layers of the network is quite generic and is useful regardless of any particular classification problem [12]. In the extreme case, only the last classification layer is retrained resulting in significantly reduced training size requirement, computation time, and hardware specifications.

One of the many available pre-trained image classifiers is the Inception v3 that works with the Python Tensorflow library. It is trained using the ImageNet 2012 dataset that contains about 14 million images and 1000 classes. This model achieves a 21.2% top-1 and a 5.6% top-5 error rate at a cost of 5.6 billion multiply-adds per inference and less than 25 million parameters [4]. Resources, including tutorials and a python script, are readily available online in retraining the last layer of Inception v3 making it more readily accessible for experimentation compared to other existing CNN image classifiers. Retraining the last layer of the Inception v3 model only takes a couple of minutes in a GPU-enabled laptop using the default settings. It has been repurposed as a flower classifier using a dataset containing 3670 images of five kinds of flowers with a 90-95% accuracy [13]. It has also been used in apparel classification to classify five apparel classes namely t-shirts, pants, saree, ladies kurta, and footwear with a per class accuracy ranging from 95.3% to 100% [5]. Inception v3 has also been used to classify forty-eight classes of consumer products ranging from baby shoes to telephones by using the features produced by the model to train an SVM classifier with a classification accuracy of 95.4% [14]. These examples show how the features extracted by the Inception v3 model provide useful information in classifying images outside its intended purpose.

In this paper, we explore how the Inception v3 model can be retrained for face shape classification. We present experimental results from tests performed using 500 female celebrity face shapes collected from the Internet. We then benchmark the results against face shape classifiers based on

LDA, SVM-LIN, SVM-RBF, MLP, and KNN and trained using pre-identified facial features enumerated in [2].

## III. EXPERIMENTAL SETUP

### A. Overview

Retraining the Inception v3 model for face shape classification and then benchmarking it with other face shape classifiers required the following steps: making a face shape dataset, extracting features needed by the non-CNN classifiers, training the classifiers, and finally benchmarking the model accuracies.

### B. Dataset collection

Several human face datasets are available online [15]. However, these datasets were used for various other applications such as face localization, facial landmarks detection, or face recognition. Without proper labelling of the correct face shape, these datasets are not readily usable in training face shape classifiers. Manual labelling of face shapes is especially problematic because expert judgment is required and in case of conflicting opinions, multiple opinions may be needed.

To overcome this issue, we reviewed several online articles to collate female celebrity names with a cited face shape. Five face shapes are considered namely heart, oblong, oval, round, and square. A total of 21 online sites were reviewed for a total of 263 celebrity names. Of these, 24 have unanimous citations from three or more articles, 62 have conflicting citations from two or more articles, and 177 have one or two non-conflicting citations. From these, 17 or 18 celebrities were chosen per face shape with 5 or 10 images per celebrity for a total of 100 images per face shape or a total of 500 images for the whole dataset.

The final number is essentially limited by the ease of finding images of the shortlisted celebrities that satisfy the following criteria. One of the major criteria is that the facial landmarks must be correctly identified by the facial landmark detection tools discussed in the next subsection. This means that the 19 facial landmark coordinates needed in training and testing the non-CNN based classifiers must be at the correct position when visually inspected. This constraint necessitates that the images be front facing and upright as much as possible to minimize distortions in the facial landmark coordinates. If the image is front facing but is inclined at an angle, the image is manually rotated so that it becomes upright. There should also be only one face detected per image which necessitated the cropping of extraneous parts for some images. There should be minimal hair occlusion such that the hairline and the outline of the face is visible. And since the Inception v3 model resizes any input image to make it square, white space is padded to the sides to make the images square to minimize distortion.

### C. Feature extraction

As in [2], 19 features were extracted per image to train the non-CNN based classifiers. The 19 features include the following: face height to width ratio, jawline width to face width ratio, chin-to-mouth to jaw line width ratio, and the angles that the line from the chin to each one of 16 facial points make with respect to the vertical. Angles with respect to the vertical were used (instead of horizontal) to take into account the slight inclination of the faces in some of the images. The features were then normalized such that the per feature mean is 0 and the per feature variance is 1 using Eq. 2.

$$z_{ij} = \frac{x_{ij} - \mu_j}{\sigma_j} \quad (2)$$

We used the OpenCV library [16] to detect the location of the face and determine the coordinates of the bounding box which in turn served as input to DLib's 68-point facial landmark detector [17]. Since the available facial landmark detector does not include the coordinates of the hairline that is needed to compute the face height, a custom hairline detector was designed. Pixels above the nose are checked until a pre-determined color difference from the initial point is detected. As mentioned in the previous subsection on dataset collection, only images that have correctly positioned facial landmark coordinates, verified by visual inspection, are included in the dataset.

Fig. 2 shows the relevant facial landmark points used in computing the 19 features used in training the non-CNN classifiers. Table I shows the formula used to compute the features from the facial landmark coordinates.

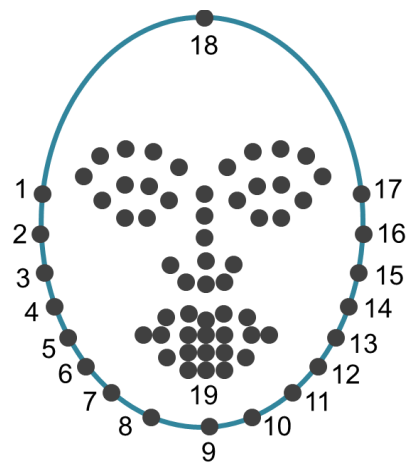

**Fig. 2** Facial landmark points for feature extraction

TABLE I.     FORMULA FOR THE 19 FEATURES EXTRACTED PER IMAGE

| Feature ID | Formula |
|---|---|
| 1 | $f_1 = \dfrac{d(9,18)}{d(1,17)}$ |
| 2 | $f_2 = \dfrac{d(5,13)}{d(1,17)}$ |
| 3 | $f_3 = \dfrac{d(9,19)}{d(5,13)}$ |
| 4-11 | $f_i = atan\left|\dfrac{x(i-3)-x(9)}{y(i-3)-y(9)}\right|; i = 4, \ldots 11$ |
| 12-19 | $f_i = atan\left|\dfrac{x(i-2)-x(9)}{y(i-2)-y(9)}\right|; i = 12, \ldots 19$ |

*D. Training the LDA, SVM, MLP, and KNN classifiers*

Python's scikit-learn 0.18.1 machine learning library's built-in codes were used to train and test classifiers using LDA, SVM, MLP, and KNN. The following discussion provides a brief description of each method and the parameters used.

Linear discriminant analysis finds a linear combination of features that separates two or more classes which can be used as a classifier or as a form of dimensionality reduction. During training for face-shape classification, we chose to reduce the high-dimensional feature vector to two components preceding classification.

Support vector machines are another type of classifier that tries to map the points in the feature space with the widest gap between classes as possible. A one-versus-one approach was used with a linear (SVM-LIN) and radial basis function (SVM-RBF) as kernels using a c-parameter of 0.01.

Artificial neural networks, also known as multi-layer perceptron, are an interconnected network of computational units called artificial neurons that try to approximate non-linear functions via a series of linear transformations and non-linear activation functions. The network used uses two hidden layers with five and two neurons in the first and second layers respectively. Parameter estimation is via a quasi-newton method called LBFGS.

Finally, k-nearest neighbour classifiers determine the class-membership of a point by looking at the training samples that are closest to it by distance, usually the Euclidean distance. The chosen model looks at the five nearest neighbors for classification.

*E. Retraining the Inception v3 classifier*

The retrain.py script was used to retrain the last layer of the Inception v3 model (https://github.com/tensorflow/tensorflow/blob/master/tensorflow/examples/image_retraining/retrain.py). Default parameters for the learning rate of 0.01 and the number of iterations of 4000 was used.

*F. Performance benchmarking*

Each classifier was trained for a training set size of 100, 200, 300, 400, and 500 images. For each training set size, the training accuracy was recorded as a measure of fit and the prediction accuracy for all 500 images was noted for overall classifier performance.

IV. EXPERIMENTAL RESULTS

Table II shows the training accuracy of each classifier as a function of the training set size while Table III shows the overall accuracy of each classifier in predicting the correct label of each of the 500 images as a function of the training set size. For both overall fit and performance, the retrained Inception v3 models outperform the other classifiers by about 20% to 30% or more. The overall accuracy of the retrained Inception v3 model ranges from 84.8% up to 97.8% while that of the other classifiers only range from 36.4% to 64.6% depending on the size of the training set used.

The overall accuracy of the retrained Inception v3 model is much greater at 84.4% to 97.8% than the reported overall accuracy in [2] using LDA, SVM-LIN, SVM-RVM, and MLP in the range of 64.2% to 70.8% using 500 images. This is also comparable to the reported accuracy in [1] using KNN in the range of 80% to 90% using 300 images.

These experimental results show that retraining the Inception v3 image classifier can significantly outperform traditional classifiers that use handpicked features in face shape recognition. However, because of the limited size of the training dataset, we can't really say much about the generalization capabilities of this model when tested on images outside the training dataset. It is therefore necessary to expand the available labeled dataset in future works so that proper model cross-validation can be performed.

TABLE II.     TRAINING ACCURACY VS. TRAINING SIZE

| Training Size | 100 | 200 | 300 | 400 | 500 |
|---|---|---|---|---|---|
| LDA | 73.0% | 63.0% | 60.7% | 62.3% | 61.6% |
| SVM-LIN | 53.0% | 55.0% | 53.0% | 52.0% | 55.2% |
| SVM-RBF | 66.0% | 57.0% | 54.0% | 53.3% | 50.6% |
| MLP | 57.0% | 60.0% | 55.7% | 54.8% | 54.0% |
| KNN | 57.0% | 64.0% | 61.7% | 59.3% | 64.6% |
| IV3* | 100.0% | 100.0% | 99.0% | 98.7% | 97.8% |

TABLE III. OVERALL ACCURACY VS. TRAINING SIZE

| Training Size | 100 | 200 | 300 | 400 | 500 |
|---|---|---|---|---|---|
| **LDA** | 52.8% | 56.4% | 58.0% | 62.2% | 61.6% |
| **SVM-LIN** | 44.4% | 48.6% | 52.8% | 53.0% | 55.2% |
| **SVM-RBF** | 45.8% | 49.0% | 50.2% | 52.2% | 50.6% |
| **MLP** | 36.4% | 46.8% | 49.6% | 52.0% | 54.0% |
| **KNN** | 43.8% | 52.0% | 55.8% | 57.6% | 64.6% |
| **IV3*** | 84.8% | 84.6% | 84.4% | 84.4% | 97.8% |

V. CONCLUSION AND FUTURE WORK

We have presented in this paper the experimental results obtained from using the retrained Inception v3 image classifier in face shape classification. As far as we know, this is the first in the published literature to use convolutional neural networks in face shape classification.

The retrained Inception v3 model has an overall accuracy ranging from 84.8% to 97.8%, outperforming classifiers based on linear discriminant analysis, support vector machines, multi-layer perceptron, and k-nearest neighbors which only had overall accuracies ranging from 36.4% to 64.6%, depending on the number of images used to train the classifiers.

However, due to the small number of training images available at only 500 labeled images, little can be said about the generalization capabilities of the model when it comes to correctly classifying images outside the this dataset. Therefore, researchers who would continue this work should consider building a much larger labeled dataset, hopefully one that can be freely distributed to the research community, so that proper model cross-validation can be performed.